\documentclass{edm_article}
\usepackage{makecell} 
\usepackage{placeins}
\usepackage{cite}
\usepackage{graphicx} 
\usepackage{booktabs} 
\usepackage{multirow} 
\usepackage{amsmath,amssymb,amsfonts,float}
\usepackage{algorithm} 
\usepackage{algpseudocode}
\usepackage{graphicx}
\usepackage{textcomp}
\usepackage{xcolor}
\usepackage{hyperref}
\usepackage{pgfplots}
\pgfplotsset{compat=1.17}
\usetikzlibrary{patterns}
\usepackage{nicematrix}
\usepackage{tikz}
\usepackage{graphicx}  
\usepackage{epstopdf} 
\usepackage{booktabs} 
\usepackage{needspace}

\usetikzlibrary{shapes.geometric, arrows, arrows.meta, positioning, calc, decorations.pathreplacing, matrix}
\usepackage{float}
\usepackage{titlesec}
\usepackage{colortbl}
\usepackage{pgfplotstable}
\usepgfplotslibrary{colormaps}
\usepackage{comment}
\usepackage{booktabs}
\usepackage{xcolor}
\def\BibTeX{{\rm B\kern-.05em{\sc i\kern-.025em b}\kern-.08em
    T\kern-.1667em\lower.7ex\hbox{E}\kern-.125emX}}
\tikzset{
    node/.style={
        circle, minimum size=0.9cm, text centered, draw=black
    },
    edge/.style={
        thick, -{Stealth}, shorten >=1pt
    },
    highlight/.style={
        rectangle, rounded corners, minimum width=2cm, minimum height=0.6cm, text centered, draw=black, fill=pastelpink
    },
    process/.style={
        rectangle, minimum width=2.5cm, minimum height=0.65cm, text centered, draw=black, fill=pastelyellow
    },
    highlight2/.style={
        rectangle, rounded corners, minimum width=2cm, minimum height=0.6cm, text centered, draw=black, fill=pastellime
    },
    dashednoarrow/.style={
        thick, dashed, -  
    }
}
\definecolor{darkgreen}{rgb}{0.0, 0.5, 0.0}  
\begin{document}

\title{Predicting Student Success with Heterogeneous Graph Deep Learning and Machine Learning Models}


\numberofauthors{1}
\author{
Anca Muresan, Mihaela Cardei, and Ionut Cardei\\
       \affaddr{Department of Electric Engineering and Computer Science}\\
        \affaddr{Florida Atlantic University}\\
       \email{ \{amuresan2023, mcardei, icardei\} @fau.edu}
}

\maketitle

\begin{abstract}
Early identification of student success is crucial for enabling timely interventions, reducing dropout rates, and promoting on-time graduation. In educational settings, AI-powered systems have become essential for predicting student performance due to their advanced analytical capabilities. However, effectively leveraging diverse student data to uncover latent and complex patterns remains a key challenge. While prior studies have explored this area, the potential of dynamic data features and multi-category entities has been largely overlooked.
To address this gap, we propose a framework that integrates heterogeneous graph deep learning models to enhance early and continuous student performance prediction, using traditional machine learning algorithms for comparison. Our approach employs a graph metapath structure and incorporates dynamic assessment features, which progressively influence the student success prediction task. 
Experiments on the Open University Learning Analytics (OULA) dataset demonstrate promising results, achieving a 68.6\% validation F1 score with only 7\% of the semester completed, and reaching up to 89.5\% near the semester’s end. Our approach outperforms top machine learning models by 4.7\% in validation F1 score during the critical early 7\% of the semester, underscoring the value of dynamic features and heterogeneous graph representations in student success prediction.

\end{abstract}


\section{Introduction}
Student success is critical for the educational sector; fortunately, the fast-paced advancements in intelligent systems have aided the adaptation of innovative strategies of predicting timely student performance \cite{b16}. In this manner, at-risk students are easily identified, and proactive measures such as personalized guidance, and immediate intervention by professionals or smart tutoring systems, can be employed to facilitate improved retention rates and overall student experience. In this study, we define student success as successfully passing the class by the end of the course.
Predicting student success is not only vital for academic institutions but also for shaping future workforce readiness, which has far-reaching societal implications.

In today's data-driven world, developing a comprehensive understanding of optimal strategies for managing and assessing vast volumes of collected data is crucial. Tracking student success early and throughout the semester is essential for effective intervention.

Higher educational institutions across the globe actively seek to develop learning methodologies that foster enhanced learning outcomes for students and ensure appropriate support for at-risk students. As a result of the significant advancements in data learning analytics and Artificial Intelligence (AI) powered technology, valuable opportunities arise to leverage existing big educational data for actionable insights. Efforts have been made to improve pedagogical practices by employing Machine Learning (ML) models for classification, regression, or clustering tasks.

Traditional ML models are limited in recoding the causal complexities inherent in the data, often treating it from a static perspective, where features do not change over time. Moreover, these methods often fail to account for the dynamic relationships in student data, which changes during the program of study, limiting their ability to provide accurate predictions throughout the semester. 

To mitigate these challenges, it is essential to model data in a way that closely reflects real-world dynamics. 
Heterogeneous graph data structures and Deep Learning (DL) models offer a promising approach for capturing complex systems, providing a framework that more closely mirrors real-world dynamics. Models such as Heterogeneous Graph Attention Network (HAN) \cite{b4}, and Heterogeneous Graph Transformer (HGT) \cite{HGT} enable the representation of intricate relationships and interactions, opening new avenues for investigating underlying patterns and enhancing our understanding of student success.
Recent efforts in researching graph DL models demonstrate their versatility and performance, as a result of capturing the subtle nuanced interactions.

This paper initially employs ML models to predict student success early in the semester. To advance this objective, we extend the approach by developing heterogeneous graph DL models, incorporating dynamic features. This allows for the integration of various node types and dynamic node features, enabling us to analyze their impact on predicting student success through a binary node classification task.

The current study is grounded in the hypothesis that incorporating dynamic features alongside existing data enhances the prediction of student success. 
For our experiments, we use the well-established Open University Learning Analytics Dataset (OULA) \cite{b3}, known for its extensive coverage and longitudinal structure.
This dataset serves as a robust test case for applying the proposed models in different configurations and addressing research questions, with the primary goal of predicting student success early in a course. The choice of OULA provides a real-world educational setting, offering a valuable opportunity to explore and validate early prediction models. 

The key contributions of this study are summarized as follows: 

\begin{itemize} \setlength\itemsep{-0.5 em}
\item 
\textbf{Implementation and adaptation of heterogeneous graph models (HAN and HGT)}: We transformed the OULA tabular dataset into heterogeneous graph datasets and explored various feature assignments, demonstrating that the introduced dynamic feature is crucial for model performance.
\item 
\textbf{Feature ablation study}: We conducted a feature ablation analysis on the heterogeneous graph datasets to assess the impact of different feature categories on predicting student success.
\item
\textbf{Comparison with traditional ML models}: We evaluated the performance of graph-based deep learning models against traditional ML models, such as Logistic Regression (LR) and Random Forest (RF).
\item
\textbf{Improved early prediction performance}: The proposed graph-based deep learning models achieved a validation F1 score of 68.6\% with only 7\% of the semester completed, representing a 4.7\% improvement over the best-performing ML models. This underscores the effectiveness of dynamic assessment features in making early and accurate student success predictions.

\item
\textbf{Educational impact}: The results highlight the potential of dynamic features and heterogeneous graph models in prediction tasks, offering educators actionable insights for timely interventions to enhance student outcomes.
\end{itemize}

These findings suggest that dynamic features and heterogeneous graph models can complement existing ML approaches, providing a pathway for improvements in early prediction systems and enabling more targeted student support.
By demonstrating how advanced learning analytics, particularly dynamic features, enhance early predictions of student success, this study contributes to the growing field of AI in education. It also underscores the potential of contextual data to inform pedagogical practices and support data-driven decision making.

This work addresses the following Research Questions (RQ):\\
RQ1: How do heterogeneous graph-based models with dynamic features and various metapath architectures compare to classical ML models in prediction tasks, particularly in the context of a sparse graph structure? \\
RQ2: What is the impact of feature selection, dynamic node features, and dataset dimensionality on the prediction of student success?\\ 
The remainder of the paper is structured as follows. Section II reviews related literature. Section III describes the dataset and outlines the methodology, including the proposed models and their implementation. Section IV presents the experimental results and analysis. Finally, Section V concludes the paper with a discussion of the findings.

\section{Related Work}
The field of Educational Data Mining (EDM) is an interdisciplinary research avenue that employs Data Mining (DM), ML, and AI techniques to assess and optimize educational processes. During the last decade, EDM has become increasingly prominent \cite{b12}, highlighting substantial impact on pedagogical strategies.
Student performance modeling leverages advanced computational techniques to predict academic outcomes by analyzing numerous factors that contribute to and influence students' academic performance, such as learning behaviors, engagement styles, demographics, and prior academic accomplishments \cite{b13}.
Several studies have explored learning patterns and quantified students' attention during various educational scenarios using neural networks \cite{b11}, and have analyzed the functional intent of students during the learning process \cite{b10}. 
Previous research has employed ML models trained on limited dataset such as midterm exams, to predict final exam grades \cite{b16}. 

Recent studies have advanced the understanding of the student performance by using the anonymized OULA dataset.
In \cite{b13}, the OULA dataset was examined and features were divided into three categories, demographics, engagement, and performance metrics, to determine the relevance of each type for student exam performance prediction. Explainable AI (XAI) models have been deployed in \cite{b14}, specifically SHapley Additive exPlanations (SHAP) in order to predict students at risk of failure, together with feature importance analysis; their models shown an improved performance in comparison with the RF baseline model. Long Short-Term Memory (LSTM) models were utilized to predict students at risk of withdrawal by leveraging the click-stream data from the OULA dataset for a binary classification task, focusing exclusively on pass and fail outcomes \cite{b18}. 
The study further examined the progression of student activity throughout the semester, starting from the 5th week, when the model achieved approximately a 60\% validation recall score.

In \cite{b15}, the authors predict student success by analyzing the initial phases of courses to their completion by utilizing convolutional  neural networks. The classic binary task is expanded to four classes (Pass, Distinction, Fail, and Withdrawn) and features are gradually considered into the model during experimentation, leading to improved predictive performance. In addition to considering a progressive semester forecasting, joint models such as Recurrent Neural Network (RNN) and Gated Recurrent Unit (GRU) have been combined in \cite{b19} to perform binary classification (pass/fail). Their results depict an accuracy of 60\% for the fifth week and 90\% for the 39th week.

A pair of graph-based neural networks was employed in \cite{b20} for binary classification, where the interaction-based graph module captured local academic representations, while the attribute-based graph module utilized graph convolutions to learn global representations from student features.
In \cite{b21}, the authors utilized the OULA dataset and proposed a Sequential Conditional Generative Adversarial Network (SC-GAN) for the synthetic generation of student records. Student performance prediction was framed as a binary classification task, and their approach outperformed traditional techniques such as synthetic minority oversampling and random oversampling, achieving an improvement of over 6\% in the area under the curve (AUC) metric.
Multi-Topology Graph Neural Networks (MTGNN) are proposed in \cite{b22}, to predict at-risk students from the OULA dataset. Their results show progressive improvements during the course of the semester, and they use two pairs of binary classification (Pass/Fail and Pass/Withdrawn).

The Graph Neural Network (GNN) model \cite{GNN} expands the capabilities of classic neural networks by directly working with graph data structures.
In \cite{b4}, the authors propose the Heterogeneous Graph Attention Network (HAN), a model that can handle heterogeneous graphs while integrating the attention mechanism in both node and semantic levels. Multiple unidirectional sequence metapath structures were deployed for evaluation.
In \cite{HGT}, a transformer-inspired model was introduced, the Heterogeneous Graph Transformer (HGT), where dynamic heterogeneity is addressed by utilizing temporal encoding to encapsulate the structure of the graph. To dynamically model inter-agent interactions, dedicated edge features was proposed in \cite{HEAT}. 
Some studies \cite{b1} examine the impact of intra-metapath and inter-metapath composition in heterogeneous graphs for recommendation systems, while others \cite{b7} focus on subgraph sampling to evaluate the significance of metapath distribution in risk detection systems. 
The use of spatial-temporal networks and synchronous dependencies between students has been explored to enhance performance prediction \cite{b9}. 

Despite significant advancements in predicting student success using ML and DL models, a research gap remains in leveraging heterogeneous graph neural networks and dynamic node features for early and continuous prediction of student performance throughout the semester.

Different than prior studies, this work utilizes dynamic features and relationships between key academic entities (such as courses, students, and registrations) to enable early academic outcome predictions through incremental feature assignment analysis. We construct heterogeneous graphs, define relevant metapaths, node types, and node features, and apply graph-based models to generate accurate and timely predictions.

\section{Dataset and Methodology}

\subsection{OULA DATASET}\label{AA}
The anonymized OULA dataset, published by Open University, a leading online institution specializing in distance learning, provides valuable insights into student engagement and performance \cite{b3}. It includes data from seven courses (referred to as modules) and spans four semesters (referred to as presentations), covering a total of 32,593 students, offering a comprehensive view of student academic life.
The dataset consists of multiple files; however, for this study, we focus exclusively on three key files, as detailed below:
\begin{itemize} \setlength\itemsep{-0.5 em}
    \item \textbf{Student Information}: This file contains details about students, including their registration status, demographic characteristics, academic background, and final results in their enrolled courses.
    \item \textbf{Course Assessments Information}: This file provides course-related details such as course code, semester, assessment type, and the weight of each assessment in the final grade.
    \item \textbf{Student Assessment}: This file includes information on student submissions for course assessments, recording assessment and student IDs, and the scores achieved.
\end{itemize}

\begin{figure*}[t]  
\Description{Image that illustrates the four steps of the proposed framework}
\centering
\includegraphics[height=0.35\textwidth]{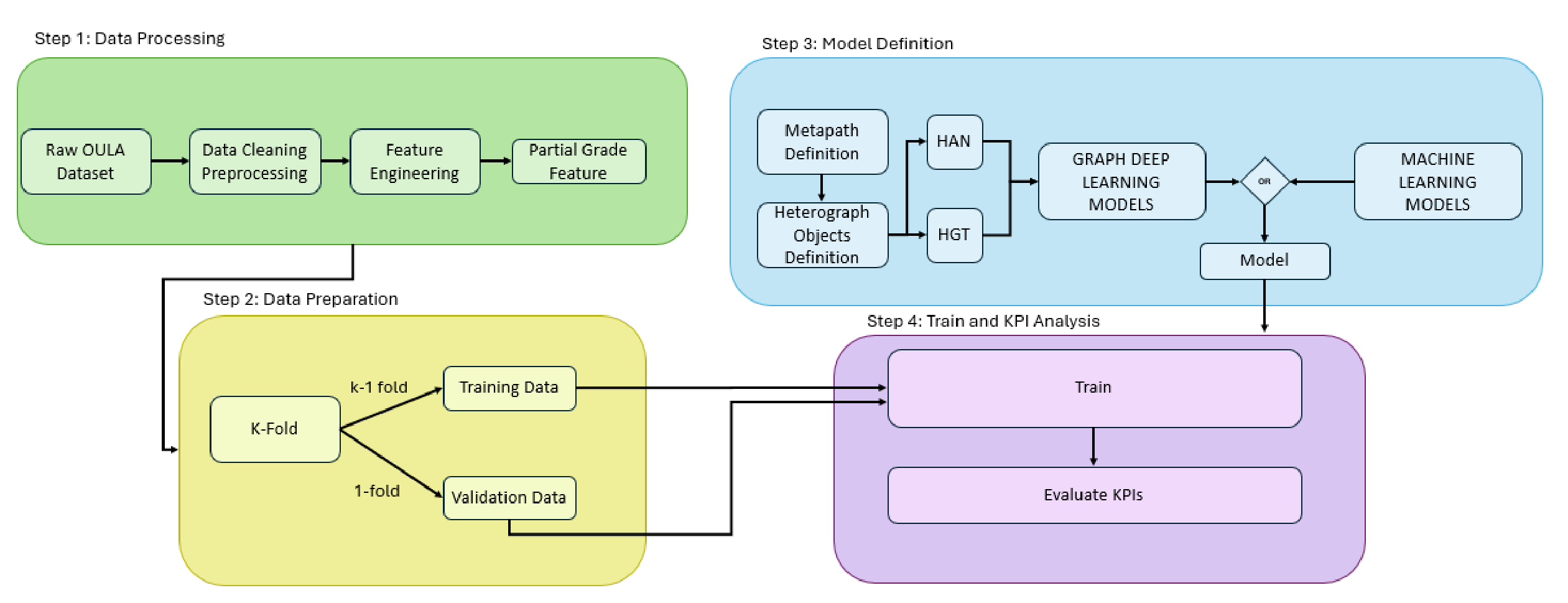}  
\caption{Proposed framework}
\label{fig:framework_correct}  
\end{figure*}

\begin{table}[t]
\renewcommand{\arraystretch}{0.6} 
\centering
\begin{small}
\caption{OULA dataset features}
\label{Features}
\begin{tabular}{@{}ll@{}}
\toprule
\textbf{Feature}       & \textbf{Description}  \\ \midrule
\multicolumn{2}{l}{\textbf{Original Student Features}} \\ \midrule
Student ID            & Index identifying the student  \\ 
Gender                & Gender of the student  \\ 
IMD Band              & Index of Multiple Deprivation band \\ 
Age Band              & Age of the student \\ 
Disability            & Student's disability status \\ 
Region                & Student's residence during the course\\ 
Education             & Highest education level \\ \midrule
\multicolumn{2}{l}{\textbf{Original Course Features}} \\ \midrule
Previous Attempts     & Number of attempts for this module\\ 
Studied Credits       & Total credits for current modules \\ 
Code Module           & Course name \\ 
Code Presentation     & Enrolled semester code\\ 
Course Category       & STEM or Social Science category \\ 
Final Result*         & Student’s final result in the course \\ 
                      & *target variable\\ \midrule

\multicolumn{2}{l}{\textbf{Engineered Feature}} \\ \midrule
Partial Grade        & Dynamic assessment grade \\ \midrule
\textbf{Total}        & \textbf{13 Original + 1 Engineered Feature} \\ \bottomrule
\end{tabular}
\end{small}
\end{table}
\subsection{METHODOLOGY}
The proposed research framework, shown in Figure \ref{fig:framework_correct}, comprises four main steps: (1) data processing, (2) data preparation, (3) model definition, and (4) training and analysis of Key Performance Indicators (KPIs).

\subsubsection{Data Processing and Preparation}
In this section, we describe the first two steps of the proposed framework.
The data processing step integrates the three aforementioned OULA files into a unified dataset, creating a cohesive representation of student data for further analysis. The process begins with the raw Student Information file, followed by an initial filtering stage. The first criterion involves excluding records for courses without available grades, ensuring the dataset’s integrity and relevance.

The original dataset categorizes Final Result into four possible outcomes: Distinction, Pass, Fail, and Withdrawn. To remove outliers, records where the Final Result was inconsistent with the dataset’s predefined grading threshold were excluded. Specifically, the dataset defines a passing grade as 40 or above and a failing grade as below 40. Additionally, students labeled as ‘Withdrawn’ who had withdrawn from the course before the semester started or within the first week were also removed.

For this study, since students classified as Withdrawn have a final grade below 40 and those awarded Distinction have a grade above 40, we retained only two categories: Pass (label 1) and Fail (label 0). Specifically, Pass and Distinction were merged into a single class (label 1), while Fail and Withdrawn were combined into another class (label 0). In this context, student success is defined as having a Final Result labeled as 1, representing either a Pass or Distinction.

Students who were enrolled in at least one course, regardless of the semester, were included in the resulting dataset, referred to as the preprocessed dataset. If a student took multiple courses, they would have as many entries in the dataset as the number of courses they enrolled in across the four available semesters.
For example, if a student was enrolled in one course per semester over two different semesters, the preprocessed dataset would contain two separate entries for the same student ID, with the Final Result serving as the target variable for each entry. 

Table \ref{Features} presents a summary of all features considered in this study. The original features are categorized into two types: student features and course features. Student features capture key demographic, socioeconomic, and academic background information relevant to learning outcomes. Each student is uniquely identified by Student ID, while Gender and Age Band provide basic demographic details. IMD Band represents socioeconomic status based on residential deprivation levels, and Disability indicates whether a student has a registered disability. Region denotes the student’s location at the time of enrollment, while Education reflects the highest qualification attained before enrollment, serving as an indicator of academic preparedness. These features offer essential context for analyzing factors influencing student success.

The course features represent the structural and academic aspects of a student’s enrollment. Code Module and Code Presentation uniquely identify each course and its corresponding semester. Semester indicates the term in which the course was taken, while Studied Credits reflect the total academic workload of the modules a student is enrolled in. Final Result captures the student’s overall performance, categorized as Pass, Fail, Withdrawn, or Distinction, providing a key outcome variable for analysis.
These features help contextualize student performance within different course structures and learning environments. According to the original OULA dataset paper \cite{b3}, courses are classified into STEM or Social Sciences, as indicated by the Course Category feature.

In addition to the original features in the dataset, we introduced a supplementary engineered dynamic feature, Partial Grade, to offer a more comprehensive perspective on student academic performance throughout the semester (see Table~\ref{Features}). A feature is considered dynamic when its values change over time, reflecting variations in student progress and engagement during the course.

To determine the Partial Grade at different points in the semester, we employed a LR model to compute the weights ($\alpha$ and $\beta$) for assignments and exams available up to those dates.

For labels, we used the Final Result feature, where students with a final grade of 40 or above were labeled as Pass, and those below 40 as Fail. The original OULA dataset includes the Final Result, indicating a student’s course outcome. However, while a passing grade corresponds to 40 or above, the dataset documentation does not specify how the final grade is computed from individual assessment grades and the final exam score.

Since the final grade directly determines the Final Result (Pass or Fail) and the formula for its calculation is not provided in the dataset, along with missing Exam Grades for several courses, we developed a logistic regression-based model to compute the $\alpha$ and $\beta$ weights for each course and semester.
There are two types of grades: assessment grades, received incrementally throughout the semester from assignments, and the 
final exam grade, awarded at the end of the semester.

In Equation~\ref{eq:final_grade}, $x$ represents the student's weighted assessment grade at the end of the semester, and $y$ represents the final exam grade. The $\alpha$ and $\beta$ values are computed using logistic regression ensuring that the equation’s output aligns with the Pass/Fail classification label.
\begin{equation}
\alpha \cdot x + \beta \cdot y \geq 40
\label{eq:final_grade}
\end{equation}
 
The model results indicate that the Exam Grade contributes significantly (approximately 90\%) to the Final Grade. Given that the target variable is the Final Result and the exam is taken at the end of the semester, we excluded the Exam Grade from the preprocessed dataset to prevent introducing bias into the model.

The dynamic engineered feature, Partial Grade, is derived from the evolution of student performance and is calculated at 13 points throughout the semester based on assessment grades. Since assessments completed near the end of the semester contribute minimally to the final grade, their influence on the overall result is relatively small. The remaining percentage is attributed to the Exam Grade, which was excluded from the preprocessed dataset to prevent bias.

These Partial Grades are computed at specific intervals of course completion, providing a dynamic reflection of student performance over time, as detailed in Table~\ref{tab:course_completion}. By excluding the Exam Grade, the Partial Grade focuses solely on assessments completed within the semester, measured at specific time thresholds (e.g., the first 20 days, first 40 days, etc.), effectively capturing incremental progress in student performance.
Although assessments contribute a maximum of about 10\% to the final grade, this feature is leveraged to make early predictions about student outcomes.

This approach highlights the potential of leveraging early-semester student performance data to predict the Final Result and offer valuable insights into student progress throughout the course.

\begin{table*}[t]
\centering
\begin{small}
\caption{Partial Grade computed at specified percentages of the course completion}
\label{tab:course_completion}
\begin{tabular}{c|c|c|c|c|c|c|c|c|c|c|c|c|c}
\toprule
\multicolumn{1}{c|}{\textbf{Percentage of Semester Completion}} & \textbf{7\%} & \textbf{15\%} & \textbf{23\%} & \textbf{30\%} & \textbf{38\%} & \textbf{46\%} & \textbf{54\%} & \textbf{60\%} & \textbf{70\%} & \textbf{77\%} & \textbf{85\%} & \textbf{93\%} & \textbf{100\%} \\
\midrule
\multicolumn{1}{c|}{\textbf{Days Since Semester Start}} & \textbf{20} & \textbf{40} & \textbf{60} & \textbf{80} & \textbf{100} & \textbf{120} & \textbf{140} & \textbf{160} & \textbf{180} & \textbf{200} & \textbf{220} & \textbf{240} & \textbf{260} \\
\bottomrule
\end{tabular}
\end{small}
\end{table*}

After preprocessing, cleaning, and filtering the original dataset, the refined dataset consists of 21,305 unique students, corresponding to 24,615 entries. For the binary classification task, the dataset was divided into two classes: Fail (Class 0), representing 43.65\% of the entries, and Pass (Class 1), accounting for 56.35\% of the entries. This distribution provides a balanced and representative sample for classification analysis.
The final phase in preparing the dataset for use within the framework involved categorical feature encoding. To ensure compatibility with the model, both label encoding and one-hot encoding were applied. Specifically, one-hot encoding was used for categorical features such as Education, Region, Code Module, and Code Presentation, while the remaining categorical variables were label encoded.

\begin{figure}[t]
\Description{Image with the principal component analysis heatmap}
\centering
\includegraphics[width=0.5\textwidth]{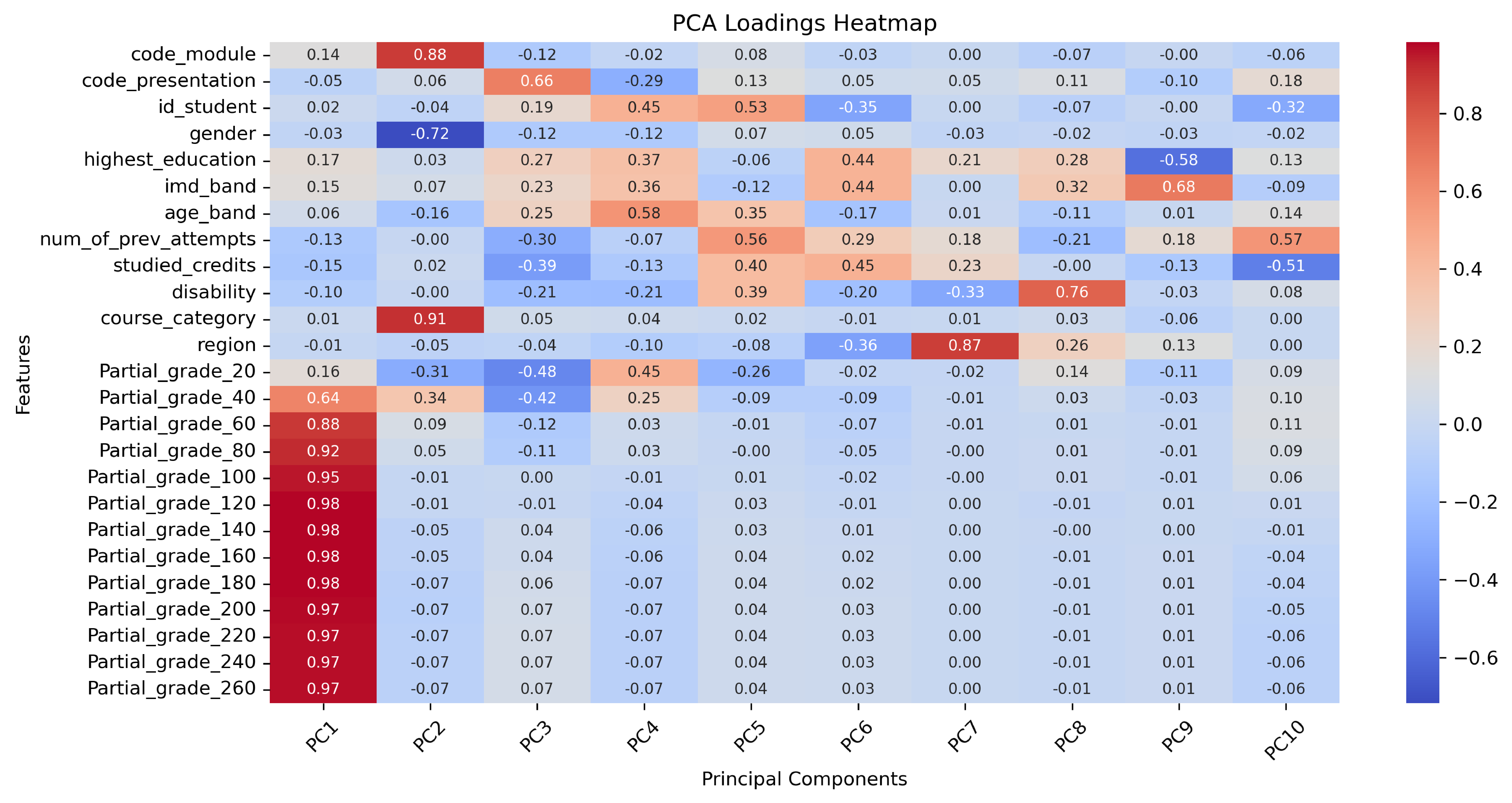} 
\caption{Principal Component Analysis (PCA) for assessing the impact of features}
\label{fig:PCA_impact}
\end{figure}

To evaluate the impact of various features on student success prediction, we conducted a feature analysis using Principal Component Analysis (PCA). The results, presented in Figure \ref{fig:PCA_impact}, illustrate the influence of key features on overall data variability and student performance.
PCA is a dimensionality reduction technique that simplifies datasets by reducing the number of variables while preserving as much of the original variation as possible \cite{PCA}. In our dataset, which includes features related to student demographics, academic details, and assessment performance, the partial grade features play a particularly crucial role. These features represent students' cumulative partial grades at specific points throughout the semester. Consequently, partial grade features significantly shape the principal components, underscoring their importance in understanding student success.

In the second step of our framework, data preparation, we apply 5-fold cross-validation (k=5) to ensure a reliable model evaluation. The dataset is divided into five folds, where four are used for training and one for validation. This process is repeated five times, with each fold serving as the validation set once. By cycling through all subsets, we ensure a comprehensive and consistent model evaluation, improving the accuracy of performance assessment.
At the end of data preparation, we obtain the training and validation datasets, which are used for model training and evaluation. To capture student progress over time, we transformed the original dataset into 13 identical datasets, each corresponding to a different prediction time interval within the semester. The only difference between these datasets is the Partial Grade feature, which dynamically changes to reflect the prediction time in the semester.

\subsubsection{Model Definition}
The third step of our framework is model definition, where we develop two types of models.
First, we establish a set of baseline models using machine learning (ML) techniques applied to tabular data. We explore nine distinct models, categorized based on their underlying principles, to identify the most effective approaches for our analysis:

\begin{itemize} \setlength\itemsep{-0.5 em}
    \item \textbf{Linear models}: Logistic Regression (LR) and Linear Discriminant Analysis (LDA), known for their efficiency in classification tasks.
    \item \textbf{Tree-based models}: Random Forest (RF) \cite{RF} and Decision Tree \cite{DecisionTree}, capable of capturing complex decision boundaries.
    \item \textbf{Instance-based learning}: K-Nearest Neighbors (KNN) \cite{KNN}, which classifies based on similarity measures.
    \item \textbf{Probabilistic models}: Quadratic Discriminant Analysis (QDA) \cite{QDA} and Gaussian Naïve Bayes \cite{GaussianNB}, leveraging probability distributions for classification.
    \item \textbf{Non-linear models}: Support Vector Classifier (SVC) \cite{SVC}, which handles complex decision boundaries using the RBF kernel.
    \item \textbf{Ensemble methods}: Bagging Classifier \cite{Bagging}, which enhances performance through model aggregation.
\end{itemize}
These models were selected for their interpretability and strong performance on structured datasets. They serve as a reliable benchmark for evaluating the effectiveness of more advanced models, enabling us to establish a baseline before exploring more complex architectures.

Secondly, we apply advanced graph-based models, specifically the Heterogeneous Graph Attention Network (HAN) \cite{b4} and the Heterogeneous Graph Transformer (HGT) \cite{HGT}. Unlike the baseline models, these approaches utilize heterogeneous graph structures to capture complex relationships between different data types. HAN employs both node-level and semantic-level attention, allowing it to focus on relevant features at multiple levels of the graph, while HGT leverages transformer mechanisms to efficiently model time-varying dependencies in large datasets. To facilitate this analysis, we construct two heterogeneous graph objects, one for training and one for validation. The creation of these graph objects involves defining node types, assigning appropriate node features, establishing edge connectivity between nodes, and determining edge directionality. This study explores the critical role of integrating multiple node types and assigning relevant node features in advanced graph-based DL models. Specifically, we examine a symmetric metapath architecture to assess its impact on both model performance and interpretability.

The proposed metapath architecture follows a registration - student - registration (RSR) structure, incorporating two distinct node types: student (S) nodes and registration (R) nodes. This structure captures both individual student characteristics and course-specific attributes, allowing for a more comprehensive representation of the learning environment. In this setup, the transitive relationships R - S and S - R in the original heterogeneous graph are summarized into an indirect metapath R - R, linking two registrations that belong to the same student.

Student nodes encode demographic and academic attributes that are essential for understanding student engagement and performance. These attributes include region, age band, IMD band, gender, disability status, number of studied credits, and highest level of education. By incorporating these features, the model can capture variations in student backgrounds and assess their potential impact on academic success.

Registration nodes serve as dynamic representations of students' academic progress. To evaluate the impact of different feature sets, we conduct a feature selection process using an ablation study. Initially, the registration node includes only the partial grade as an evolving feature, reflecting student performance throughout the semester. In subsequent configurations, additional attributes are introduced to enhance the contextual richness of the model. Specifically, the second configuration incorporates the number of previous attempts, while later cases extend the feature set by including course-related attributes such as course category, code module, and code presentation. Additionally, the registration node contains the target variable, the Final Result (Pass/Fail), which serves as the basis for binary node classification.

To systematically assess the impact of different features, we propose five distinct cases as part of the ablation study, as detailed in Table~\ref{FeatureAssignment}. Case 1 includes only the temporal grade, serving as the baseline. Case 2 extends this representation by integrating the number of previous attempts, capturing the influence of prior engagement with the course. Case 3 introduces course-related attributes, incorporating a broader context of the learning environment. Case 4 further enhances the model by adding student-specific attributes, including highest education level attained, number of studied credits, and gender. Finally, Case 5 represents the most comprehensive configuration, encompassing all available student demographic features, including disability status, region, age, and IMD band.
The feature ablation study demonstrates that accurate predictions can be achieved without relying on personal data such as age, gender, education level, disability status, IMD band, or previously studied credits. This highlights the potential for privacy-preserving methods in academic performance monitoring.
For HGT, using Case 3 features instead of Case 5 results in no more than a 1.9\% decrease in F1 score and only a 0.9\% drop in accuracy, even when predicting outcomes 20 days into the semester.

By gradually expanding the feature set, this study evaluates the contribution of each attribute category to predictive performance, providing insights into the role of dynamic and heterogeneous data representations in modeling student success. Upon completing the graph creation stage, two heterogeneous graph objects are generated, based on unique students, their corresponding courses, and selected node features. These graphs serve as input for the advanced models during the training and validation phases.

\begin{table}[t]
\centering
\begin{small}
\renewcommand{\arraystretch}{0.5} 
\caption{Feature ablation study cases}
\label{FeatureAssignment}
\begin{tabular}{@{}lccccc@{}}
\toprule
\midrule
\textbf{Feature} & \textbf{Case 1} & \textbf{Case 2} & \textbf{Case 3} & \textbf{Case 4} & \textbf{Case 5} \\ \midrule
\multicolumn{6}{l}{\textbf{Registration node features}} \\ \midrule
{\makecell[l]{Partial \\ Grade}} & \checkmark & \checkmark & \checkmark & \checkmark & \checkmark \\ \midrule
{\makecell[l]{Previous \\ Attempts}} &  & \checkmark & \checkmark & \checkmark & \checkmark \\ \midrule 
{\makecell[l]{Course \\ Category}} &  &  & \checkmark & \checkmark & \checkmark \\ \midrule
{\makecell[l]{Code \\ Module}} &  &  & \checkmark & \checkmark & \checkmark \\ \midrule
{\makecell[l]{Code \\ Presentation}} &  &  & \checkmark & \checkmark & \checkmark \\ \midrule
\multicolumn{6}{l}{\textbf{Student node features}} \\
\midrule
{\makecell[l]{Education}} &  &  &  & \checkmark &  \checkmark \\ \midrule
{\makecell[l]{Studied \\ Credits}} &  &  &  & \checkmark & \checkmark \\ \midrule   
{\makecell[l]{Gender}} &  &  &  &  \checkmark & \checkmark \\ \midrule
{\makecell[l]{Disability}} &  &  &  &  & \checkmark \\ \midrule
{\makecell[l]{Region}} &  &  &  &  & \checkmark \\ \midrule
{\makecell[l]{Age}} &  &  &  &  & \checkmark \\ \midrule
{\makecell[l]{IMD Band}} &  &  &  &  & \checkmark \\ \midrule
\bottomrule
\end{tabular}
\end{small}
\end{table}

\subsubsection{Train and Key Performance Indicators Analysis}
The fourth step in the framework is training and performance evaluation. Once the model is selected, training is conducted using the training dataset, while the validation dataset is used to monitor performance. To enhance model performance and reduce overfitting, we implement early stopping and fine-tune the model’s parameters through hyperparameter tuning. The model’s effectiveness is then evaluated using key performance metrics, including Training F1 Score and Accuracy as well as Validation F1 Score and Accuracy.

\begin{table*}[t] 
\centering
\caption{Empirical evaluation of ML models performance}
\label{table:all_baseline}
\renewcommand{\arraystretch}{0.9} 
\begin{small}
\begin{tabular}{@{}llrrrrrrrrrrrrr@{}} 
\toprule
\multicolumn{2}{c}{} & \multicolumn{13}{c}{\textbf{Days}} \\ 
\cmidrule(lr){3-15}
\textbf{Model} & \textbf{Metric} & \textbf{20} & \textbf{40} & \textbf{60} & \textbf{80} & \textbf{100} & \textbf{120} & \textbf{140} & \textbf{160} & \textbf{180} & \textbf{200} & \textbf{220} & \textbf{240} & \textbf{260} \\ \midrule
\midrule
\multirow{4}{*}{\makecell[l]{Logistic \\ Regression}} & Train Accuracy & 0.648 & 0.681 & 0.757 & 0.772 & 0.815 & 0.835 & 0.854 & 0.870 & 0.878 & 0.883 & 0.889 & 0.889 & 0.889 \\  
 & Train F1 & 0.640 & 0.674 & 0.751 & 0.768 & 0.813 & 0.833 & 0.853 & 0.870 & 0.877 & 0.882 & 0.889 & 0.889 & 0.889 \\  
 & Val Accuracy & \textbf{0.648} & 0.679 & \textbf{0.757} & \textbf{0.772} & \textbf{0.815} & \textbf{0.835} & \textbf{0.854} & 0.869 & 0.876 & 0.882 & 0.889 & 0.889 & 0.889 \\  
 & Val F1 & \textbf{0.639} & 0.672 & \textbf{0.750} & \textbf{0.767} & \textbf{0.812} & \textbf{0.833} & \textbf{0.853} & 0.868 & 0.876 & 0.881 & 0.889 & 0.889 & 0.889 \\ \midrule

\multirow{4}{*}{LDA} & Train Accuracy & 0.648 & 0.681 & 0.754 & 0.769 & 0.811 & 0.833 & 0.853 & 0.870 & 0.880 & 0.885 & 0.892 & 0.892 & 0.892 \\  
 & Train F1 & 0.639 & 0.673 & 0.744 & 0.761 & 0.806 & 0.830 & 0.851 & 0.869 & 0.878 & 0.885 & 0.892 & 0.892 & 0.892 \\  
 & Val Accuracy & 0.647 & 0.679 & 0.754 & 0.769 & 0.810 & 0.833 & 0.853 & 0.870 & \textbf{0.879} & \textbf{0.885} & 0.892 & 0.892 & 0.892 \\  
 & Val F1 & 0.638 & 0.671 & 0.744 & 0.761 & 0.806 & 0.830 & 0.851 & 0.868 & \textbf{0.878} & \textbf{0.884} & 0.891 & 0.891 & 0.891 \\ \midrule

\multirow{4}{*}{\makecell[l]{Random \\ Forest \\ Classifier}} & Train Accuracy & 0.967 & 0.989 & 0.999 & 0.999 & 1.000 & 1.000 & 1.000 & 1.000 & 1.000 & 1.000 & 1.000 & 1.000 & 1.000 \\  
 & Train F1 & 0.967 & 0.989 & 0.999 & 0.999 & 1.000 & 1.000 & 1.000 & 1.000 & 1.000 & 1.000 & 1.000 & 1.000 & 1.000 \\  
 & Val Accuracy & 0.637 & \textbf{0.686} & 0.746 & 0.765 & 0.811 & 0.831 & 0.852 & \textbf{0.870} & \textbf{0.879} & 0.884 & \textbf{0.892} & \textbf{0.892} & \textbf{0.892} \\  
 & Val F1 & 0.634 & \textbf{0.683} & 0.743 & 0.762 & 0.809 & 0.829 & 0.851 & \textbf{0.869} & \textbf{0.878} & 0.884 & \textbf{0.892} & \textbf{0.892} & \textbf{0.892} \\ \midrule

\multirow{4}{*}{\makecell[l]{KNN \\ Classifier}} & Train Accuracy & 0.745 & 0.771 & 0.809 & 0.821 & 0.849 & 0.866 & 0.883 & 0.896 & 0.904 & 0.909 & 0.917 & 0.917 & 0.917 \\  
 & Train F1 & 0.742 & 0.768 & 0.807 & 0.819 & 0.848 & 0.865 & 0.882 & 0.895 & 0.903 & 0.909 & 0.917 & 0.917 & 0.917 \\  
 & Val Accuracy & 0.620 & 0.664 & 0.730 & 0.744 & 0.791 & 0.818 & 0.845 & 0.862 & 0.871 & 0.877 & 0.885 & 0.885 & 0.885 \\  
 & Val F1 & 0.615 & 0.660 & 0.726 & 0.741 & 0.789 & 0.816 & 0.844 & 0.861 & 0.870 & 0.876 & 0.885 & 0.885 & 0.885 \\ \midrule

\multirow{4}{*}{\makecell[l]{Gaussian \\ Naive \\ Bayes}} & Train Accuracy & 0.613 & 0.628 & 0.703 & 0.724 & 0.767 & 0.783 & 0.810 & 0.828 & 0.839 & 0.846 & 0.852 & 0.852 & 0.852 \\  
 & Train F1 & 0.615 & 0.629 & 0.702 & 0.724 & 0.767 & 0.783 & 0.810 & 0.828 & 0.839 & 0.846 & 0.852 & 0.852 & 0.852 \\  
 & Val Accuracy & 0.610 & 0.623 & 0.701 & 0.723 & 0.767 & 0.781 & 0.808 & 0.827 & 0.838 & 0.845 & 0.851 & 0.851 & 0.851 \\  
 & Val F1 & 0.611 & 0.624 & 0.701 & 0.723 & 0.767 & 0.781 & 0.808 & 0.827 & 0.838 & 0.845 & 0.851 & 0.851 & 0.851 \\ \midrule

\multirow{4}{*}{QDA} & Train Accuracy & 0.621 & 0.664 & 0.721 & 0.748 & 0.790 & 0.806 & 0.828 & 0.845 & 0.855 & 0.861 & 0.867 & 0.867 & 0.867 \\  
 & Train F1 & 0.622 & 0.659 & 0.704 & 0.736 & 0.784 & 0.802 & 0.825 & 0.842 & 0.853 & 0.860 & 0.866 & 0.866 & 0.866 \\  
 & Val Accuracy & 0.611 & 0.660 & 0.719 & 0.745 & 0.787 & 0.803 & 0.825 & 0.842 & 0.853 & 0.859 & 0.865 & 0.865 & 0.865 \\  
 & Val F1 & 0.611 & 0.655 & 0.702 & 0.733 & 0.780 & 0.798 & 0.822 & 0.840 & 0.852 & 0.858 & 0.864 & 0.864 & 0.864 \\ \midrule

\multirow{4}{*}{\makecell[l]{Bagging \\ Classifier}} & Train Accuracy & 0.925 & 0.949 & 0.963 & 0.966 & 0.971 & 0.974 & 0.977 & 0.980 & 0.982 & 0.982 & 0.984 & 0.984 & 0.984 \\  
 & Train F1 & 0.925 & 0.949 & 0.963 & 0.966 & 0.971 & 0.974 & 0.977 & 0.980 & 0.982 & 0.982 & 0.984 & 0.984 & 0.984 \\  
 & Val Accuracy & 0.611 & 0.653 & 0.715 & 0.734 & 0.782 & 0.806 & 0.831 & 0.849 & 0.860 & 0.867 & 0.877 & 0.877 & 0.877 \\  
 & Val F1 & 0.611 & 0.652 & 0.713 & 0.733 & 0.781 & 0.805 & 0.831 & 0.849 & 0.859 & 0.866 & 0.877 & 0.877 & 0.877 \\ \midrule

\multirow{4}{*}{SVC} & Train Accuracy & 0.631 & 0.668 & 0.743 & 0.756 & 0.804 & 0.831 & 0.850 & 0.867 & 0.878 & 0.885 & 0.892 & 0.892 & 0.892 \\  
 & Train F1 & 0.609 & 0.655 & 0.728 & 0.742 & 0.797 & 0.828 & 0.847 & 0.865 & 0.877 & 0.884 & 0.891 & 0.891 & 0.891 \\  
 & Val Accuracy & 0.629 & 0.667 & 0.742 & 0.756 & 0.803 & 0.831 & 0.849 & 0.867 & 0.877 & \textbf{0.885} & 0.892 & 0.892 & 0.892 \\  
 & Val F1 & 0.608 & 0.655 & 0.728 & 0.742 & 0.797 & 0.828 & 0.846 & 0.865 & 0.876 & \textbf{0.884} & 0.891 & 0.891 & 0.891 \\ \midrule

\multirow{4}{*}{\makecell[l]{Decision \\ Tree \\ Classifier}} & Train Accuracy & 0.967 & 0.989 & 0.999 & 0.999 & 1.000 & 1.000 & 1.000 & 1.000 & 1.000 & 1.000 & 1.000 & 1.000 & 1.000 \\  
 & Train F1 & 0.967 & 0.989 & 0.999 & 0.999 & 1.000 & 1.000 & 1.000 & 1.000 & 1.000 & 1.000 & 1.000 & 1.000 & 1.000 \\  
 & Val Accuracy & 0.585 & 0.618 & 0.675 & 0.693 & 0.746 & 0.772 & 0.796 & 0.811 & 0.828 & 0.841 & 0.853 & 0.853 & 0.853 \\  
 & Val F1 & 0.585 & 0.619 & 0.675 & 0.693 & 0.746 & 0.772 & 0.796 & 0.811 & 0.828 & 0.841 & 0.853 & 0.853 & 0.853 \\  
\midrule
\bottomrule
\end{tabular}
\end{small}

\end{table*}

\section{EXPERIMENTAL RESULTS}
In this section, we present the experimental results, evaluating the performance of various classification models using partial grades computed at different points throughout the semester. We assess both traditional ML models and heterogeneous graph-based DL models on a binary classification task, using the Final Result feature as the target variable. For the DL models, early stopping is applied with a patience of 100 epochs, and training is conducted for a maximum of 800 epochs. The Adam optimizer is used to improve convergence speed and overall performance \cite{adam}. Additionally, we perform a grid search to optimize hyperparameters, aiming for optimal model performance.

The training heterogeneous graphs contain between 19,656 and 19,701 nodes and 25,282 to 25,587 edges, all belonging to a single node type, Registration (R). The edges represent connections between R nodes, forming a structured network. The degree distribution is skewed, with the majority of nodes having a degree of one, resulting in an average degree of 1.29. A small fraction of nodes exhibit higher degrees, with a maximum of 5. These structural characteristics underscore the sparsity and connectivity of the graph, which are crucial for further analysis.
The validation heterogeneous graphs consist of 4,870 to 4,959 nodes and 6,172 to 6,477 edges, all of which belong to the R node type. The maximum degree remains 5, with most nodes maintaining a degree of 1, indicating a sparse structure similar to the training graphs. The sparse nature of the validation graph could pose challenges for the HAN model, making it harder to identify meaningful connections. In contrast, the HGT model may still perform well by effectively capturing relationships between nodes, even in cases with fewer connections.

\begin{figure*}[t]
\Description{Image comparing the performance of LDA, Logistic Regression, Random Forest and case 5 for HAN and HGT models.}
  \centering
  \includegraphics[scale=0.45]{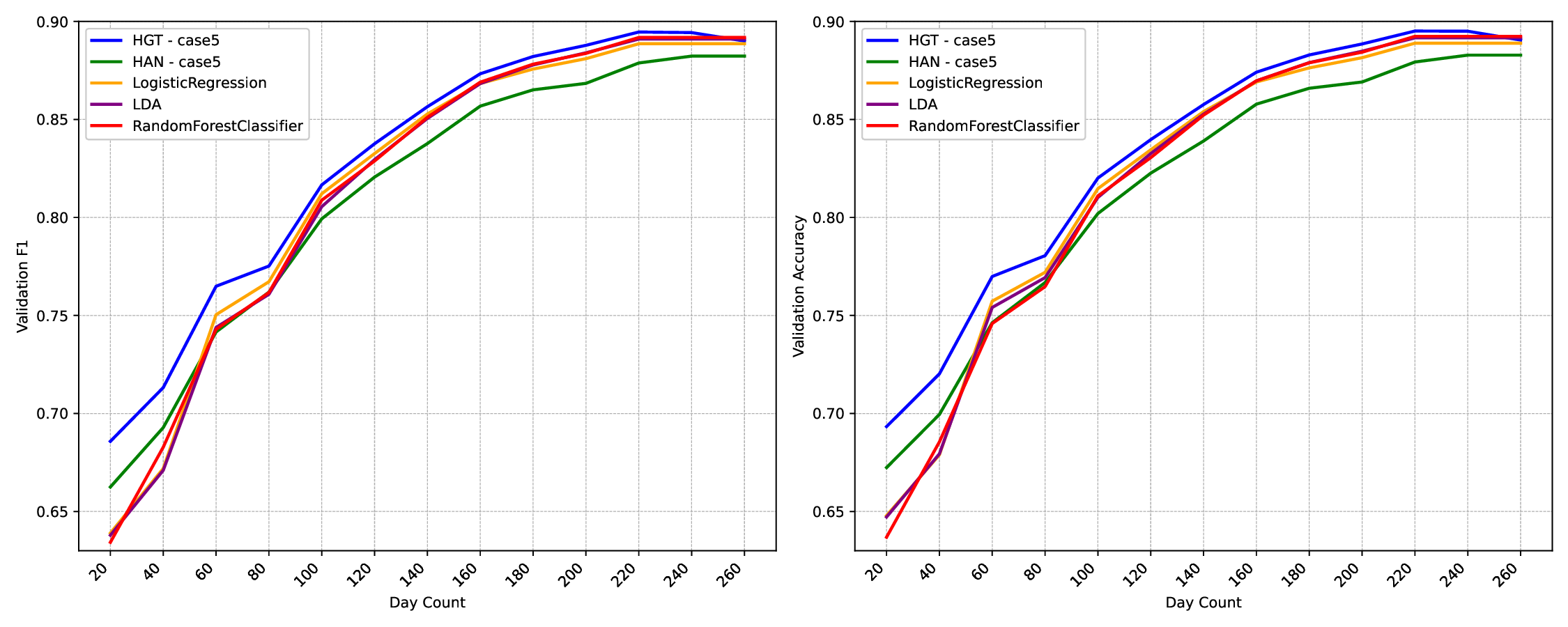}  
  \caption{Comparison of validation metrics: best performing ML models vs. case 5 graph DL models} 
  \label{fig:graphs_cases_2}  
\end{figure*}

Table~\ref{table:all_baseline} presents the performance metrics of various ML models across different time intervals, focusing on training and validation accuracy and F1 score. For these models, we used the full feature set (as defined in Case 5 in Table~\ref{FeatureAssignment}). The models, including LR and RF, were evaluated for their ability to predict student performance, with the highest values highlighted in bold. Among the models, LR and LDA consistently outperformed others in both training and validation metrics, making them the top candidates for further analysis. These models demonstrated strong predictive performance with minimal overfitting, positioning them as the most effective approaches. While RF also performs well, it exhibited signs of overfitting (see Table~\ref{table:top3vs1}). KNN provided moderate performance, whereas Gaussian Naïve Bayes and QDA showed relatively weaker predictive capabilities. Based on these findings, LR and LDA are recommended as the primary models for further analysis and potential deployment (see Figure~\ref{fig:graphs_cases_2}).

\begin{table*}[t] 
\caption{Empirical comparison of the performance of graph DL models across feature assignment cases}
\label{table:HANHGTCases}
\renewcommand{\arraystretch}{0.9} 
\centering
\begin{small}
\begin{tabular}{@{}llrrrrrrrrrrrrr@{}} 
\toprule
\multicolumn{2}{c}{} & \multicolumn{13}{c}{\textbf{Days}} \\ 
\cmidrule(lr){3-15}

\textbf{Model} & \textbf{Metric} & \textbf{20} & \textbf{40} & \textbf{60} & \textbf{80} & \textbf{100} & \textbf{120} & \textbf{140} & \textbf{160} & \textbf{180} & \textbf{200} & \textbf{220} & \textbf{240} & \textbf{260} \\ \midrule
\midrule
\multirow{4}{*}{\makecell[l]{HAN \\ Case 1}} & Train Accuracy & 0.588 & 0.651 & 0.719 & 0.742 & 0.787 & 0.810 & 0.828 & 0.846 & 0.856 & 0.879 & 0.887 & 0.886 & 0.886 \\  
 & Train F1 & 0.529 & 0.621 & 0.704 & 0.736 & 0.783 & 0.809 & 0.827 & 0.845 & 0.856 & 0.878 & 0.886 & 0.886 & 0.886 \\  
 & Val Accuracy & 0.591 & 0.653 & 0.719 & 0.743 & 0.786 & 0.811 & 0.827 & 0.846 & 0.857 & 0.879 & 0.886 & 0.886 & 0.886 \\  
 & Val F1 & 0.534 & 0.624 & 0.704 & 0.737 & 0.782 & 0.810 & 0.826 & 0.845 & 0.857 & 0.878 & 0.886 & 0.886 & 0.886 \\ \midrule


\multirow{4}{*}{\makecell[l]{HGT \\ Case 1}} & Train Accuracy & 0.555 & 0.667 & 0.741 & 0.760 & 0.806 & 0.830 & 0.850 & 0.867 & 0.877 & 0.884 & 0.891 & 0.891 & 0.890 \\  
 & Train F1 & 0.542 & 0.639 & 0.733 & 0.752 & 0.802 & 0.827 & 0.848 & 0.866 & 0.876 & 0.884 & 0.890 & 0.890 & 0.890 \\  
 & Val Accuracy & 0.555 & 0.667 & 0.743 & 0.761 & 0.807 & 0.831 & 0.851 & 0.868 & 0.877 & 0.885 & 0.891 & 0.891 & 0.892 \\  
 & Val F1 & 0.543 & 0.643 & 0.735 & 0.753 & 0.803 & 0.828 & 0.850 & 0.867 & 0.877 & 0.885 & 0.891 & 0.890 & 0.891 \\  \midrule


\multirow{4}{*}{\makecell[l]{HAN \\ Case 2}} & Train Accuracy & 0.641 & 0.661 & 0.733 & 0.750 & 0.792 & 0.817 & 0.836 & 0.855 & 0.865 & 0.881 & 0.886 & 0.887  & 0.887 \\  
 & Train F1 & 0.602 & 0.662 & 0.726 & 0.743 & 0.789 & 0.815 & 0.834 & 0.854 & 0.865 & 0.880 & 0.885 & 0.886 & 0.886 \\  
 & Val Accuracy & 0.640 & 0.662 & 0.733 & 0.749 & 0.791 & 0.818 & 0.836 & 0.854 & 0.866 & 0.881 & 0.886 & 0.887 & 0.887\\  
 & Val F1 & 0.601 & 0.663 & 0.726 & 0.742 & 0.788 & 0.816 & 0.835 & 0.853 & 0.865 & 0.880 & 0.886 & 0.886 & 0.886 \\  \midrule

\multirow{4}{*}{\makecell[l]{HGT \\ Case 2}} & Train Accuracy & 0.649 & 0.699 & 0.749 & 0.766 & 0.807 & 0.831 & 0.851 & 0.868 & 0.878 & 0.885 & 0.890 & 0.890 & 0.890 \\  
 & Train F1 & 0.621 & 0.687 & 0.743 & 0.760 & 0.804 & 0.828 & 0.850 & 0.867 & 0.877 & 0.884 & 0.890 & 0.890 & 0.890 \\  
 & Val Accuracy & 0.649 & 0.695 & 0.749 & 0.768 & 0.807 & 0.832 & 0.852 & 0.868 & 0.878 & 0.885 & 0.891 & 0.891 & 0.891 \\  
 & Val F1 & 0.622 & 0.683 & 0.743 & 0.763 & 0.804 & 0.830 & 0.850 & 0.867 & 0.877 & 0.885 & 0.890 & 0.890 & 0.890 \\  \midrule


\multirow{4}{*}{\makecell[l]{HAN \\ Case 3}} & Train Accuracy & 0.672 & 0.699 & 0.754 & 0.770 & 0.807 & 0.823 & 0.847 & 0.862 & 0.871 & 0.878 & 0.883 & 0.885 & 0.886 \\  
 & Train F1 & 0.652 & 0.691 & 0.748 & 0.764 & 0.803 & 0.821 & 0.845 & 0.861 & 0.870 & 0.877 & 0.882 & 0.884 & 0.885 \\  
 & Val Accuracy & 0.672 & 0.698 & 0.754 & 0.769 & 0.807 & 0.824 & 0.847 & 0.861 & 0.871 & 0.877 & 0.883 & 0.884 & 0.885 \\  
 & Val F1 & 0.652 & 0.690 & 0.747 & 0.763 & 0.803 & 0.822 & 0.845 & 0.860 & 0.870 & 0.877 & 0.883 & 0.884 & 0.885 \\ \midrule

\multirow{4}{*}{\makecell[l]{HGT \\ Case 3}} & Train Accuracy & 0.684 & 0.725 & 0.777 & 0.789 & 0.819 & 0.844 & 0.861 & 0.877 & 0.883 & 0.888 & 0.895 & 0.895 & 0.895 \\  
 & Train F1 & 0.669 & 0.716 & 0.771 & 0.783 & 0.816 & 0.842 & 0.859 & 0.876 & 0.882 & 0.888 & 0.895 & 0.895 & 0.895 \\  
 & Val Accuracy & 0.684 & 0.723 & \textbf{0.774} & \textbf{0.787} & 0.819 & \textbf{0.843} & \textbf{0.861} & \textbf{0.877} & \textbf{0.885} & \textbf{0.888} & \textbf{0.895} & \textbf{0.895} & 0.895 \\  
 & Val F1 & 0.668 & 0.713 & \textbf{0.768} & \textbf{0.781} & 0.815 & \textbf{0.841} & \textbf{0.859} & \textbf{0.876} & \textbf{0.884} & \textbf{0.888} & \textbf{0.895} & \textbf{0.895} & 0.895 \\ \midrule


\multirow{4}{*}{\makecell[l]{HAN \\ Case 4}} 
& Train Accuracy & 0.672 & 0.703 & 0.750 & 0.765 & 0.805 & 0.822 & 0.842 & 0.860 & 0.867 & 0.875 & 0.883 & 0.883 & 0.883 \\  
& Train F1 & 0.661 & 0.697 & 0.744 & 0.759 & 0.802 & 0.821 & 0.841 & 0.859 & 0.866 & 0.875 & 0.883 & 0.883 & 0.883 \\  
& Val Accuracy & 0.667 & 0.700 & 0.746 & 0.763 & 0.802 & 0.821 & 0.841 & 0.858 & 0.867 & 0.872 & 0.882 & 0.882 & 0.882 \\  
& Val F1 & 0.657 & 0.694 & 0.740 & 0.757 & 0.799 & 0.819 & 0.839 & 0.857 & 0.866 & 0.871 & 0.881 & 0.881 & 0.881 \\ \midrule

\multirow{4}{*}{\makecell[l]{HGT \\ Case 4}} 
& Train Accuracy & 0.706 & 0.730 & 0.775 & 0.787 & 0.820 & 0.843 & 0.861 & 0.876 & 0.883 & 0.888 & 0.896 & 0.894 & 0.896 \\  
& Train F1 & 0.699 & 0.725 & 0.770 & 0.781 & 0.817 & 0.841 & 0.860 & 0.876 & 0.883 & 0.888 & 0.896 & 0.894 & 0.896 \\  
& Val Accuracy & \textbf{0.693} & \textbf{0.725} & 0.771 & 0.782 & 0.819 & 0.839 & 0.858 & 0.874 & 0.882 & 0.887 & \textbf{0.895} & \textbf{0.895} & \textbf{0.896} \\  
& Val F1 & \textbf{0.686} &\textbf{0.720} & 0.766 & 0.776 & 0.816 & 0.837 & 0.858 & 0.873 & 0.882 & 0.887 & \textbf{0.895} & \textbf{0.895} & \textbf{0.896} \\ \midrule


\multirow{4}{*}{\makecell[l]{HAN \\ Case 5}} & Train Accuracy & 0.679 & 0.706 & 0.750 & 0.768 & 0.805 & 0.823 & 0.841 & 0.858 & 0.868 & 0.870 & 0.880 & 0.883 & 0.883 \\  
 & Train F1 & 0.669 & 0.700 & 0.745 & 0.763 & 0.802 & 0.821 & 0.840 & 0.857 & 0.867 & 0.870 & 0.880 & 0.883 & 0.883 \\  
 & Val Accuracy & 0.672 & 0.700 & 0.746 & 0.767 & 0.802 & 0.823 & 0.839 & 0.858 & 0.866 & 0.869 & 0.879 & 0.883 & 0.883 \\  
 & Val F1 & 0.663 & 0.693 & 0.742 & 0.762 & 0.799 & 0.821 & 0.838 & 0.857 & 0.865 & 0.868 & 0.879 & 0.882 & 0.882 \\ \midrule
\multirow{4}{*}{\makecell[l]{HGT \\ Case 5}} & Train Accuracy & 0.703 & 0.730 & 0.775 & 0.783 & 0.822 & 0.842 & 0.860 & 0.875 & 0.886 & 0.890 & 0.897 & 0.896 & 0.890 \\  
 & Train F1 & 0.696 & 0.723 & 0.770 & 0.778 & 0.819 & 0.840 & 0.859 & 0.874 & 0.885 & 0.889 & 0.897 & 0.896 & 0.890 \\  
 & Val Accuracy & \textbf{0.693} & 0.720 & 0.770 & 0.781 & \textbf{0.820} & 0.840 & 0.858 & 0.874 & 0.883 & \textbf{0.888} & \textbf{0.895} & \textbf{0.895} & 0.891 \\  
 & Val F1 & \textbf{0.686} & 0.713 & 0.765 & 0.775 & \textbf{0.817} & 0.838 & 0.857 & 0.873 & 0.882 & \textbf{0.888} &\textbf{0.895} & \textbf{0.895} & 0.890 \\ \midrule

\bottomrule
\end{tabular}
\end{small}
\end{table*}

\begin{figure*}[t]
    \centering
    \begin{minipage}{\textwidth}
    \Description{Image showing the performance of the five cases in the feature ablation study for HAN and HGT models}
        \centering
        \includegraphics[scale=0.45]{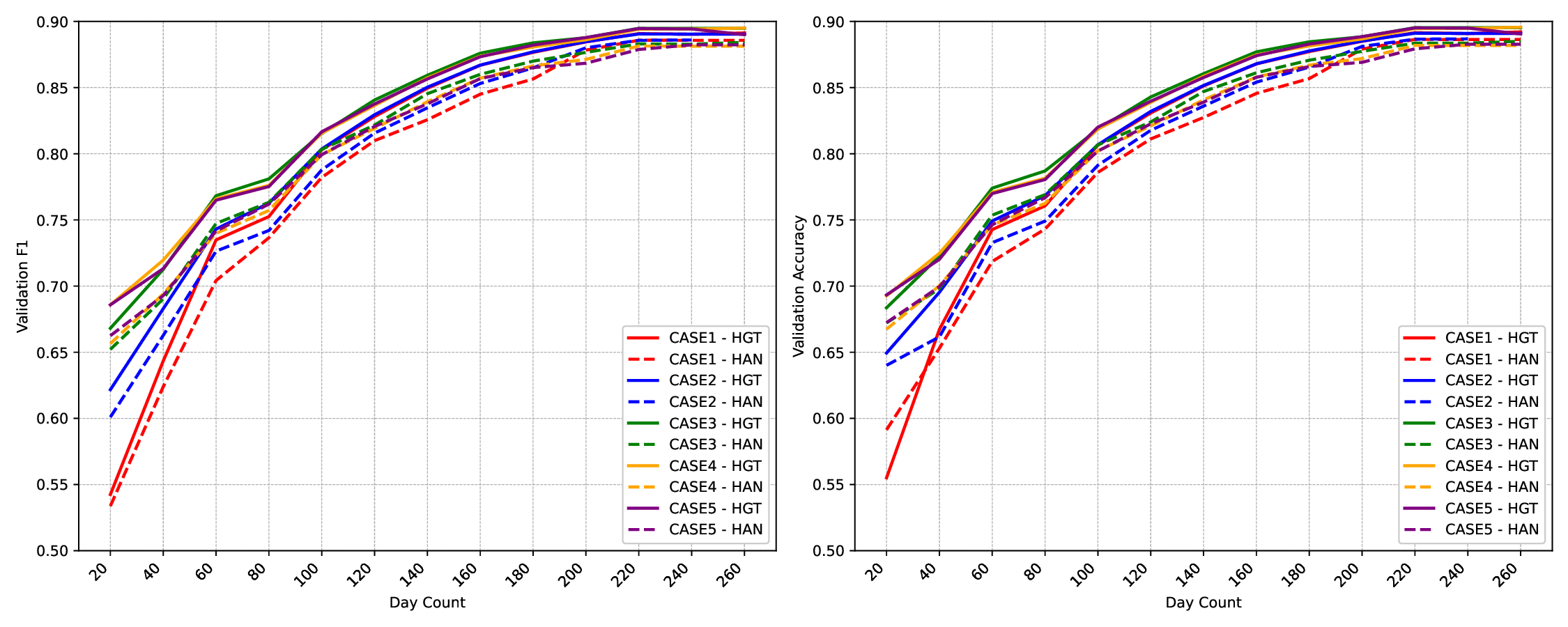}  
        \caption{Comparison of validation metrics across all cases in the feature ablation study} 
        \label{fig:graphs_cases}  
    \end{minipage}

    \begin{minipage}{\textwidth}
    \Description{Image showing the performance of case 2 and case 5 from the feature ablation study for HAN and HGT models}
        \centering
        \includegraphics[scale=0.45]{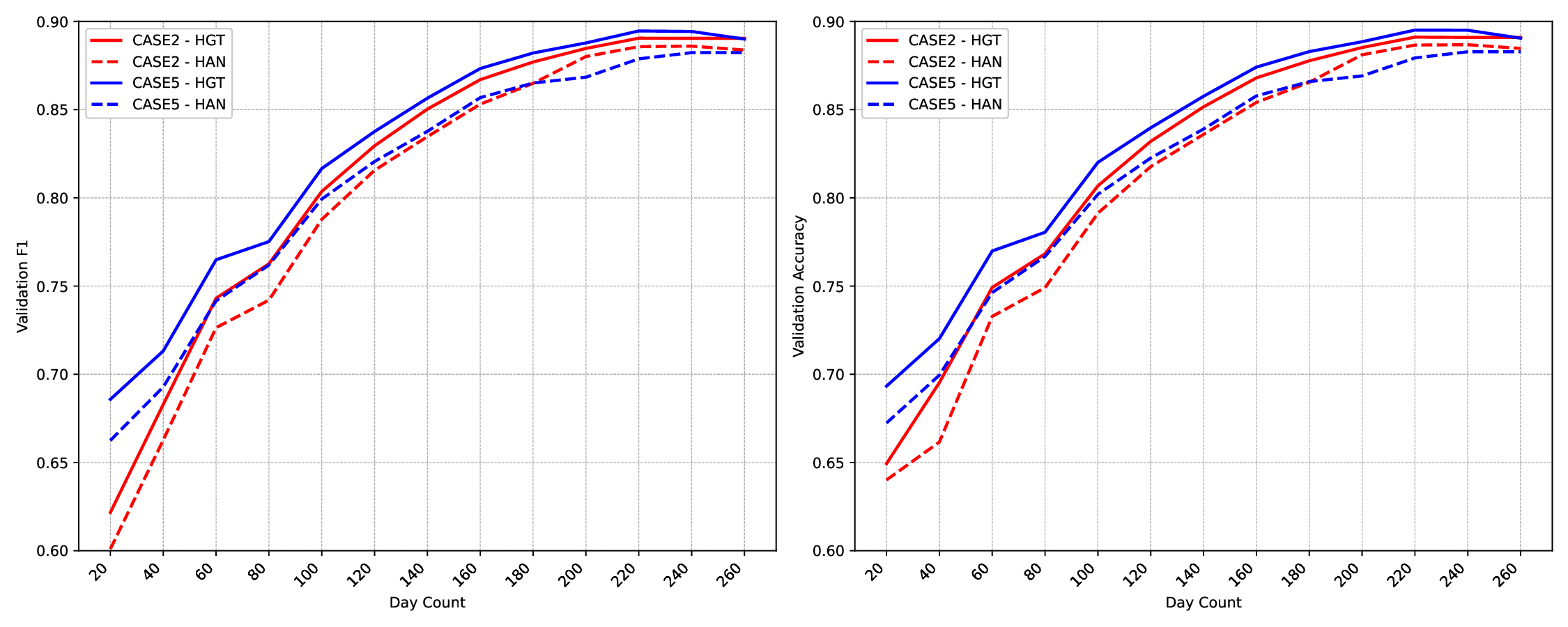}  
        \caption{Comparison of validation metrics for case 2 vs. case 5 graph DL models} 
        \label{fig:graphs_cases_3}  
    \end{minipage}
\end{figure*}

 For the heterogeneous graph DL models (HGT and HAN), we conduct a feature ablation study to evaluate the effectiveness of different feature categories. Table~\ref{table:HANHGTCases} presents the training and validation performance metrics across all cases, with the best results per time interval highlighted in bold.
Notably, the last three cases exhibit similar performance, suggesting that adding more features does not necessarily lead to substantial improvements. This aligns with the PCA analysis (Figure~\ref{fig:PCA_impact}) indicating that certain features have limited influence on final results. For example, although Case 5 incorporates additional features beyond Case 4, these additions do not enhance model performance as expected.
Figure~\ref{fig:graphs_cases} provides a comparative analysis of key performance metrics, including validation F1 score and accuracy, across all cases for HGT and HAN models. Interestingly, Case 2, which uses only two features, demonstrates unexpectedly strong performance. This suggests that even with a limited feature set, the model can achieve notable results, see Figure~\ref{fig:graphs_cases_3}. While subsequent cases introduce more features and show gradual performance improvements, these gains are not as significant as anticipated. This finding highlights a key insight: adding more features does not always result in proportional improvements in model performance.

This insight is particularly valuable in scenarios where data availability is limited, yet maintaining high model performance remains a priority. The strong results observed in Case 2 suggest that focusing on a smaller set of features can still lead to effective predictions. Validation F1 scores early in the semester (up to 40 days) indicate that course and semester information (one-hot encoded in Case 4) provide greater improvements than later in the semester, where the partial grade feature becomes a much stronger predictor of success. This suggests that early-semester predictions do not significantly benefit from additional student demographic information.

Based on the results, we conclude that Case 5 achieves the highest overall performance. Therefore, we select this model for further comparison with the top two ML models. With these insights, we now address the research questions, examining how heterogeneous graph-based models compare to traditional ML approaches, and evaluating the role of feature selection, dynamic node features, and dataset dimensionality in student success prediction.

RQ1:How do heterogeneous graph-based models with dynamic features and various metapath architectures compare to classical ML models in prediction tasks, particularly in the context of a sparse graph structure? 

RQ1 Answer: In our experiments, we compared heterogeneous graph DL models (HGT and HAN) with traditional ML models for predicting student success in a sparse graph structure. The sparsity of the graph, where nodes have few connections, posed a challenge for graph-based models. However, despite this limitation, the graph models outperformed classical ML approaches, particularly early in the semester. This suggests that even with limited connectivity, leveraging dynamic features and the relational structure of heterogeneous graphs provides valuable insights that traditional ML models may not fully capture.
Notably, HGT achieved an F1 score that exceeded the best traditional ML model by 4.7\% within the first 20 days of the semester. However, this performance gap gradually decreased to 0.4\% by the end of the semester (260 days). This trend indicates that early-semester predictions benefit significantly from the relational information captured by graph models, whereas traditional ML models become more competitive as more assessment data becomes available.

\begin{table*}[t] 
\centering
\caption{Empirical comparison of the performance of best-performing ML models and graph DL model}
\label{table:top3vs1}
\begin{small}
\begin{tabular}{@{}llrrrrrrrrrrrrr@{}} 
\toprule
\multicolumn{2}{c}{} & \multicolumn{13}{c}{\textbf{Days}} \\ 
\cmidrule(lr){3-15}
\textbf{Model} & \textbf{Metric} & \textbf{20} & \textbf{40} & \textbf{60} & \textbf{80} & \textbf{100} & \textbf{120} & \textbf{140} & \textbf{160} & \textbf{180} & \textbf{200} & \textbf{220} & \textbf{240} & \textbf{260} \\ \midrule
\midrule
\multirow{4}{*}{\makecell[l]{Logistic \\ Regression}} & Train Accuracy & 0.648 & 0.681 & 0.757 & 0.772 & 0.815 & 0.835 & 0.854 & 0.870 & 0.878 & 0.883 & 0.889 & 0.889 & 0.889 \\  
 & Train F1 & 0.640 & 0.674 & 0.751 & 0.768 & 0.813 & 0.833 & 0.853 & 0.870 & 0.877 & 0.882 & 0.889 & 0.889 & 0.889 \\  
 & Val Accuracy & 0.648 & 0.679 & 0.757 & 0.772 & 0.815 & 0.835 & 0.854 & 0.869 & 0.876 & 0.882 & 0.889 & 0.889 & 0.889 \\  
 & Val F1 & 0.639 & 0.672 & 0.750 & 0.767 & 0.812 & 0.833 & 0.853 & 0.868 & 0.876 & 0.881 & 0.889 & 0.889 & 0.889 \\ \midrule


\multirow{4}{*}{\makecell[l]{Random \\ Forest \\ Classifier}} & Train Accuracy & 0.967 & 0.989 & 0.999 & 0.999 & 1.000 & 1.000 & 1.000 & 1.000 & 1.000 & 1.000 & 1.000 & 1.000 & 1.000 \\  
 & Train F1 & 0.967 & 0.989 & 0.999 & 0.999 & 1.000 & 1.000 & 1.000 & 1.000 & 1.000 & 1.000 & 1.000 & 1.000 & 1.000 \\  
 & Val Accuracy & 0.637 & 0.686 & 0.746 & 0.765 & 0.811 & 0.831 & 0.852 & 0.870 & 0.879 & 0.884 & 0.892 & 0.892 & \textbf{0.892} \\  
 & Val F1 & 0.634 & 0.683 & 0.743 & 0.762 & 0.809 & 0.829 & 0.851 & 0.869 & 0.878 & 0.884 & 0.892 & 0.892 & \textbf{0.892} \\ \midrule
\multirow{4}{*}{\makecell[l]{HGT \\ Case 5}} & Train Accuracy & 0.703 & 0.730 & 0.775 & 0.783 & 0.822 & 0.842 & 0.860 & 0.875 & 0.886 & 0.890 & 0.897 & 0.896 & 0.890 \\  
 & Train F1 & 0.696 & 0.723 & 0.770 & 0.778 & 0.819 & 0.840 & 0.859 & 0.874 & 0.885 & 0.889 & 0.897 & 0.896 & 0.890 \\  
 & Val Accuracy & \textbf{0.693} & \textbf{0.720} & \textbf{0.770} & \textbf{0.781} & \textbf{0.820} & \textbf{0.840} & \textbf{0.858} & \textbf{0.874} & \textbf{0.883} & \textbf{0.888} & \textbf{0.895} & \textbf{0.895} & 0.891 \\  
 & Val F1 & \textbf{0.686} & \textbf{0.713} & \textbf{0.765} & \textbf{0.775} & \textbf{0.817} & \textbf{0.838} & \textbf{0.857} & \textbf{0.873} & \textbf{0.882} & \textbf{0.888} &\textbf{0.895} & \textbf{0.895} & 0.890 \\

\midrule
\bottomrule
\end{tabular}
\end{small}
\end{table*}

\begin{table*}[!ht] 
\caption{Empirical comparison of the Running Time (RT) in seconds  of best-performing ML and graph DL models}
\label{table:RT}
\centering
\begin{small}
\begin{tabular}{@{}llrrrrrrrrrrrr@{}} 
\toprule
\multicolumn{1}{c}{} & \multicolumn{13}{c}{\textbf{Days}} \\ 
\textbf{Model} & \textbf{20} & \textbf{40} & \textbf{60} & \textbf{80} & \textbf{100} & \textbf{120} & \textbf{140} & \textbf{160} & \textbf{180} & \textbf{200} & \textbf{220} & \textbf{240} & \textbf{260} \\ \midrule
\midrule
\multirow{1}{*}{LR}  & 0.365 & 0.340 & 0.344 & 0.372 & 0.330 & 0.366 & 0.360 & 0.343 & 0.400 & 0.369 & 0.400 & 0.370 & 0.353 \\  \midrule

\multirow{1}{*}{RF} & 2.001 & 1.934 & 1.856 & 1.912 & 1.728 & 1.946 & 1.884 & 1.653 & 1.575 & 1.553 & 1.506 & 1.540 & 1.522 \\ \midrule

\multirow{1}{*}{HGT Case5}  & 414.5 & 682.7 & 665.2 & 691.5 & 641.1 & 950.4 & 949.0 & 927.9 & 917.8 & 949.1 & 975.2 & 997.5 & 946.4 \\  \midrule
\multirow{1}{*}{HAN Case5}  & 159.3 & 171.5 & 175.1 & 171.3 & 171.5 & 170.0 & 171.6 & 174.3 & 170.8 & 171.0 & 141.0 & 171.6 & 170.4 \\  

\midrule
\bottomrule
\end{tabular}
\end{small}
\end{table*}

RQ2: What is the impact of feature selection, dynamic node features, and dataset dimensionality on the prediction of student success? 

RQ2 Answer: Our findings demonstrate that feature selection and dynamic node features play a critical role in predicting student success. While adding more features increases model complexity, the most meaningful performance improvements came from incorporating dynamic features that evolve over time, such as partial grades and previous attempts. Despite the increase in dataset dimensionality, adding too many features did not always lead to significant gains, especially when certain features did not meaningfully contribute to final predictions.
Thus, selecting the right set of dynamic features is essential for achieving optimal early prediction accuracy. Our results suggest that student demographic information does not have as significant an impact as one might expect. This insight is particularly relevant in contexts where privacy regulations may limit access to demographic data. Additionally, these findings indicate that our model remains robust even in the presence of stale or outdated demographic information, further emphasizing the importance of dynamic academic features over static personal attributes.

Table~\ref{table:RT} presents the Running Time (RT) performance of the best performing ML and graph DL models. While traditional models (LR, RF) offer faster execution due to their simpler architectures, the higher RT of HAN and HGT is expected as these models involve multi-layer attention mechanisms, type-specific transformations, and neighborhood aggregation across graph structures. The HGT model, in particular, achieves the best overall validation F1 score however is the most computationally intensive due to its handling of heterogeneous graphs with greater structural complexity.

\section{Conclusions}
This research highlights the effectiveness of heterogeneous graph DL models for student success prediction, particularly when leveraging dynamic features within sparse graph structures. Our approach outperformed traditional ML models by 4.7\% in validation F1 score during the early semester period, demonstrating the advantages of graph-based representations in capturing student performance patterns.
Additionally, our findings indicate that incorporating student demographic features does not always lead to significant performance improvements, as simpler models with carefully selected dynamic features can still achieve strong predictive accuracy.
Overall, this study underscores the potential of graph-based models for student success prediction, enabling timely interventions and ultimately contributing to improved academic outcomes.

%
\bibliographystyle{abbrv}
\bibliography{sigproc}  

\end{document}